\documentclass[10pt,twocolumn,letterpaper]{article}

\usepackage[pagenumbers]{cvpr} 

\usepackage{multirow}
\usepackage{array}
\usepackage{mathrsfs}

\usepackage[dvipsnames]{xcolor}

\definecolor{cvprblue}{rgb}{0.21,0.49,0.74}
\usepackage[pagebackref,breaklinks,colorlinks,citecolor=cvprblue]{hyperref}

\def\paperID{*****} 
\def\confName{CVPR}
\def\confYear{2024}

\newcommand{\model}{\textsc{Dream Engine}}

\title{Multimodal Representation Alignment for Image Generation: \\Text-Image Interleaved Control Is Easier Than You Think} 

\author{
Liang Chen\textsuperscript{1} \quad
Shuai Bai\textsuperscript{2} \quad
Wenhao Chai\textsuperscript{3} \quad
Weichu Xie\textsuperscript{4} \quad
Haozhe Zhao\textsuperscript{1} \quad
\\
Leon Vinci \textsuperscript{5} \quad
Junyang Lin\textsuperscript{2} \quad
Baobao Chang\textsuperscript{1} \quad
\\ 
[2mm]
\textsuperscript{1}~Peking University \quad \textsuperscript{2}~Alibaba Group \quad \textsuperscript{3}~University of Washington\\
\textsuperscript{4}~Beijing Institute of Technology \quad 
\textsuperscript{5}~Bainance Labs \\
[2mm]
\url{https://github.com/chenllliang/DreamEngine} 
}

\begin{document}

\twocolumn[{%
\renewcommand\twocolumn[1][]{#1}%
\maketitle
\centering
\includegraphics[width=1\linewidth]{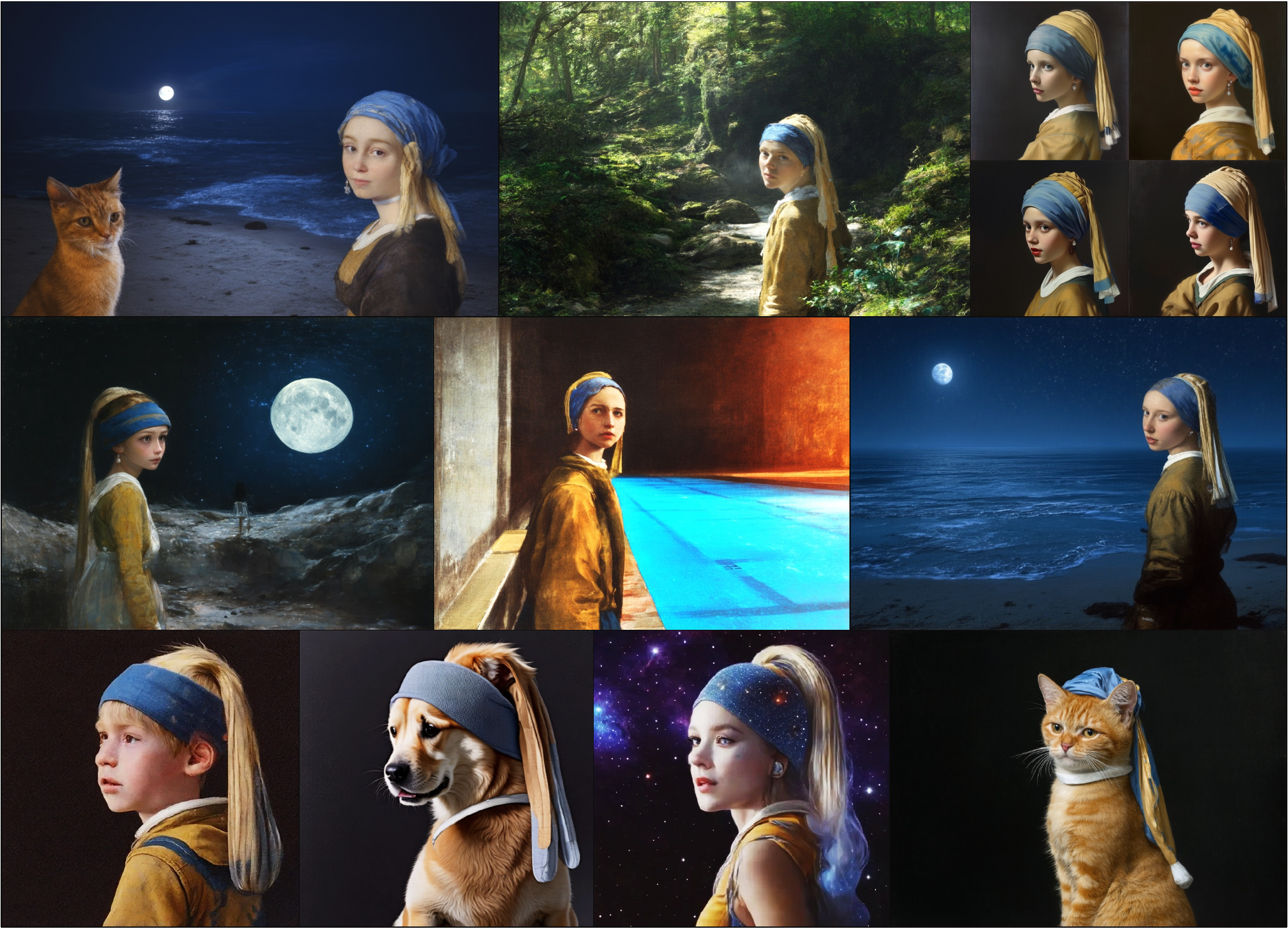}
\captionof{figure}{Generation examples of \model. Leveraging powerful text-to-image diffusion model and large multimodal models, \model~is capable of generating image with text-image interleaved control by merging concepts from different images. \vspace{2em}}
\label{fig:teaser}
}]

\clearpage

\begin{abstract}
\vspace{-1em}
The field of advanced text-to-image generation is witnessing the emergence of unified frameworks that integrate powerful text encoders, such as CLIP and T5, with Diffusion Transformer backbones. Although there have been efforts to control output images with additional conditions, like canny and depth map, a comprehensive framework for arbitrary text-image interleaved control is still lacking. This gap is especially evident when attempting to merge concepts or visual elements from multiple images in the generation process. To mitigate the gap, we conducted  preliminary experiments showing that large multimodal models (LMMs) offer an effective shared representation space, where image and text can be well-aligned to serve as a condition for external diffusion models. Based on this discovery, we propose~\model, an efficient and unified framework designed for arbitrary text-image interleaved control in image generation models. Building on powerful text-to-image models like SD3.5, we replace the original text-only encoders by incorporating versatile multimodal information encoders such as QwenVL. Our approach utilizes a two-stage training paradigm, consisting of joint text-image alignment and multimodal interleaved instruction tuning. Our experiments demonstrate that this training method is effective, achieving a 0.69 overall score on the GenEval benchmark, and matching the performance of state-of-the-art text-to-image models like SD3.5 and FLUX. 

\end{abstract}

\section{Introduction}
\label{sec:intro}

Recent years have witnessed remarkable advancements in text-to-image generation, primarily driven by powerful diffusion models~\citep{2020DDPM,ldm,2024SD3,flux}. While these models excel at generating images that align with simple text prompts, they struggle to handle more complex instructions that interweave graphical and textual elements. Although condition augmentation methods like IP-Adapter~\citep{ye2023ip-adapter} and ControlNet~\citep{controlnet} enhance text-to-image models with additional low-level control signals such as canny edges, depth maps, or reference images, they lack the flexibility to process complex and high-level text-image interleaved instructions, for example, merging visual elements from multiple images using natural language descriptions. This inability restricts more creative image generation processes where users might want to precisely orchestrate visual compositions by combining and manipulating elements from multiple sources with simple text instructions.

\begin{figure}[t]
\centering
\includegraphics[width=1\linewidth]{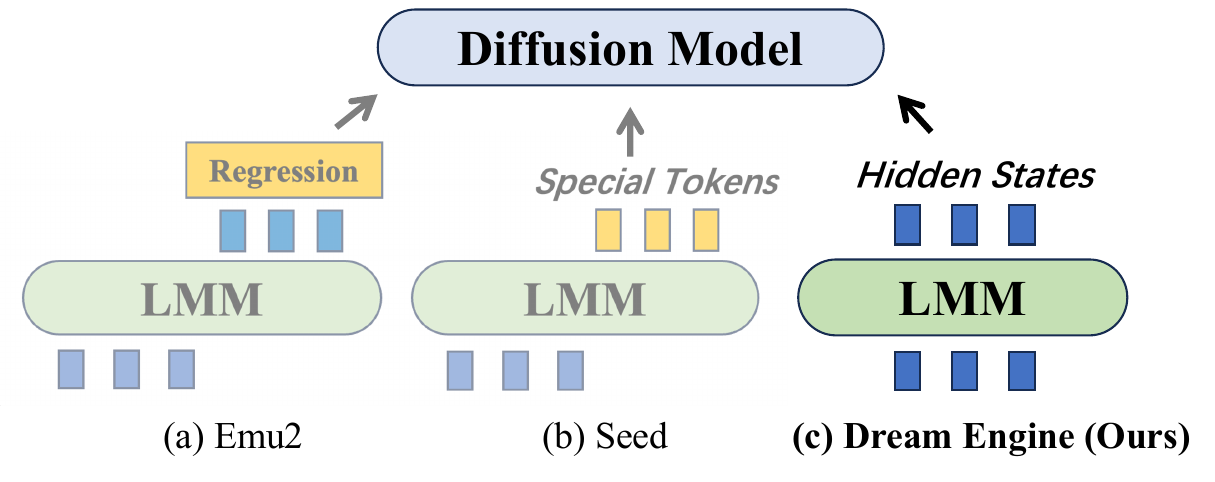}
\caption{\textbf{Overview Comparison.} Among all types of works connecting LMM and diffusion model, our~\model~adopts the simplest design yet achieves the best performance.}
\label{fig:models architecture}
\end{figure}

Meanwhile, Large Multimodal Models (LMMs)~\citep{chen2024tokenpredictionmultimodalintelligence,liu2023llava,QwenVL,Qwen2vl} have shown remarkable progress in understanding visual content and natural language instructions, enabling various tasks such as image captioning, visual question answering, and visual grounding. This advancement raises an intriguing question: \textit{Can we take advantage of the advanced visual language understanding capabilities of LMMs to improve diffusion-based image generation models, enabling more flexible text-image interleaved control?}

Several recent works have explored integrating LMMs with diffusion models to enhance image generation control. As shown in Figure~\ref{fig:models architecture}, Emu-1 and 2~\citep{sun2023emu1, emu2} incorporate a specialized regression head on the hidden output states of LMM tokens following multimodal input processing. Seed-Tokenizer~\citep{seed-tokenizer} expands the LMM vocabulary with discrete vision tokens, which serve as condition for the diffusion model during image generation. BLIP-Diffusion~\citep{li2023blipdiffusionpretrainedsubjectrepresentation} employs a multimodal query-transformer encoder to extract subject representations, which are then combined with text prompts to guide the generation process.

However, many of these approaches merely add text-to-image generation capabilities to LMMs without improving generation quality~\citep{zhao2024bridging} or expanding potential applications. Some methods~\citep{li2023blipdiffusionpretrainedsubjectrepresentation} are designed for specific tasks and can only process a single conditioning image, limiting their utility in scenarios involving multiple image inputs. To the best of our knowledge, no existing models can effectively perform compositional image generation tasks, indicating a gap in understanding text-image interleaved control, particularly when multiple images are involved.

\begin{figure*}[t]
\centering
\includegraphics[width=1\linewidth]{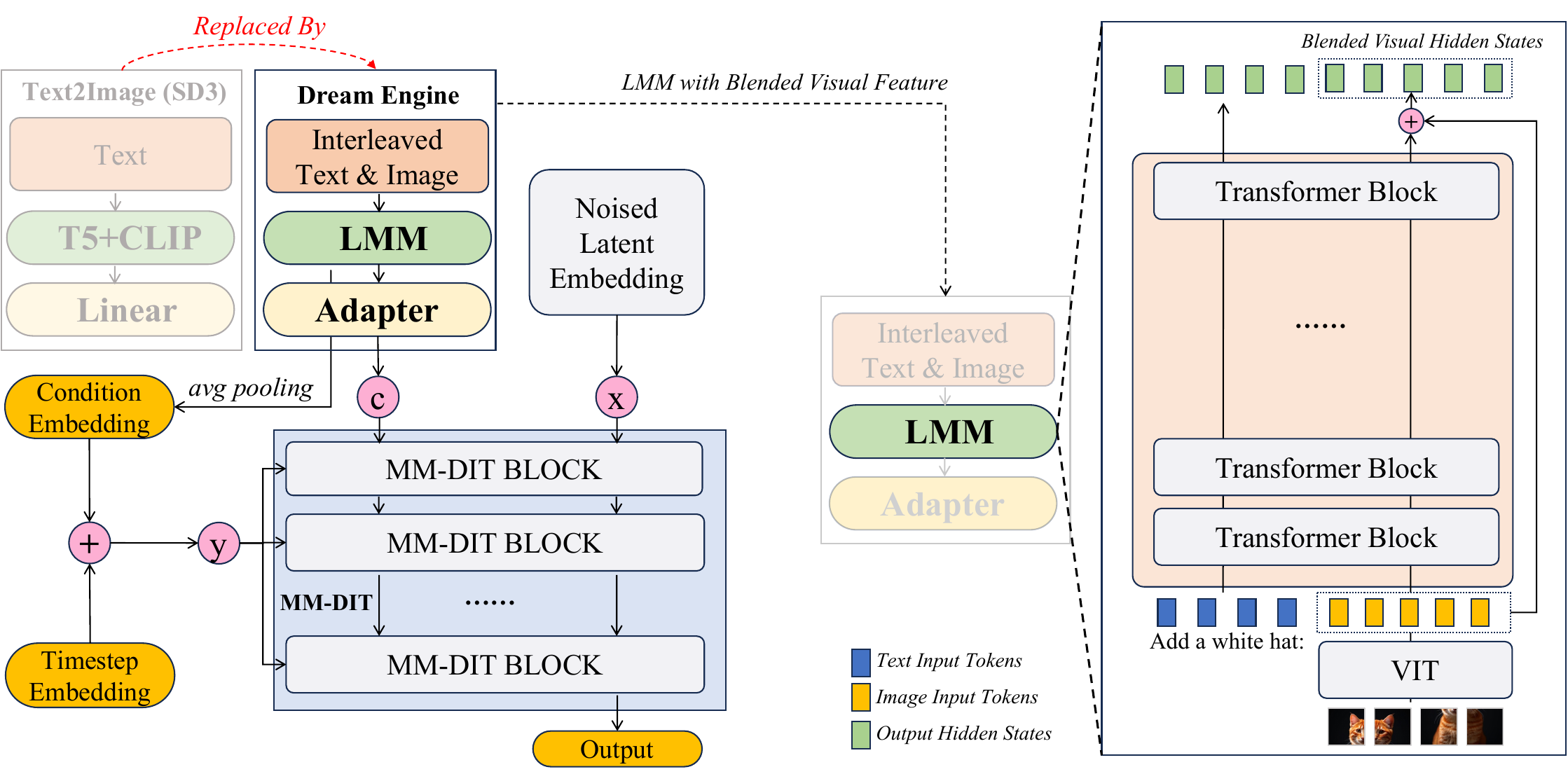}
\captionof{figure}{\textbf{\model~architecture.}}
\label{fig:dream_engine}
\end{figure*}

Our insight is that the fundamental challenge lies in effectively representing multimodal interleaved control, where mapping both text and images into a unified semantic space is crucial for coherent alignment. In this work, we demonstrate that Large Multimodal Models inherently provide a unified representation space, eliminating the need for additional architectural components such as regression heads or specialized tokens. We propose \model, an efficient and effective framework for image generation that accepts arbitrary text-image interleaved control signals. Building upon open-source text-to-image diffusion models like Stable Diffusion v3.5~\citep{2024SD3}, we replace its text encoders with a LMM along with a lightweight projector layer to encode the text-image interleaved controls. We introduce a two-stage training paradigm that efficiently aligns the representation spaces between these backbone models, enabling the generation of images guided by interleaved text and image instructions. We also design a new task called objects driven generation, which leverages object detection and image captioning data to enable compositional generation. 

Our experiments demonstrate the effectiveness of our architecture design and training recipe. By fine-tuning only an MLP layer on 20 millions data during Stage-I training, our model achieves an overall score of 0.69 on the GenEval benchmark, matching the performance of state-of-the-art text-to-image models such as SDv3.5 (0.71) and surpassing FLUX.1 Dev (0.66). This result highlights the efficacy of our alignment tuning method and demonstrates that powerful multimodal encoders can replace text encoders without compromising the original diffusion model's image generation quality. Furthermore, our Stage-II model exhibits strong text-image interleaved instruction following, significantly outperforming comparable models like Emu2-gen with substantially less training data. Notably, it can even synthesize concepts from different input images based on the text prompt to generate a cohesive output image as shown in Figure~\ref{fig:teaser}. Our contributions are threefold:

\begin{itemize}
    \item We found that Large Multi-Modal Models can be easily adapted into the text encoder of the text-to-image diffusion models even without updating the parameters.
    \item We achieve object-driven generation, combining object detection and captioning for compositional generation.
    \item Our method allows for complex, interwoven guidance from both text and images, resulting in highly customized outputs and state-of-the-art quality.
\end{itemize}

\section{Methods}
\label{sec:method}

We target at enabling diffusion models to take different text-image interleaved control in a unified manner by introducing a large multimodal model. We first concisely introduce LMM and MM-DiT architecture in Section~\ref{subsec:lmm_dit}, which are the foundational components of our method. Next, we introduce the structure design of~\model, explaining how do we align LMM with the diffusion model and how to maintain visual features consistency in Section~\ref{subsec:structure_model}. Last, we introduce the two-stage training recipe and how to curate data from different tasks in in Section~\ref{subsec:training_stages}.

\subsection{Preliminary: LMM and MM-DiT}
\label{subsec:lmm_dit}

\paragraph{Large Multimodal Models}
Large multimodal models typically include three major modules, a visual encoder usually a ViT~\citep{vit} structure, a large language model backbone and an alignment layer used to align the representation space between the visual encoder and LLM~\citep{liu2023llava,QwenVL,Qwen2vl,zhao2023mmicl,InternVL}. Given an input image \( i \), this data first passes through the visual encoder followed by the alignment layer, resulting in a transformed representation denoted as $h_{\mathrm{ViT}}$. Subsequently,   $h_{\mathrm{ViT}}$ is processed by the LLM backbone consisting of multiple transformer decoder blocks, evolving into the output image hidden states $h_{\mathrm{LLM}}$ used for text generation.

\begin{figure*}[t]
\vspace{5em}
\includegraphics[width=1\linewidth]{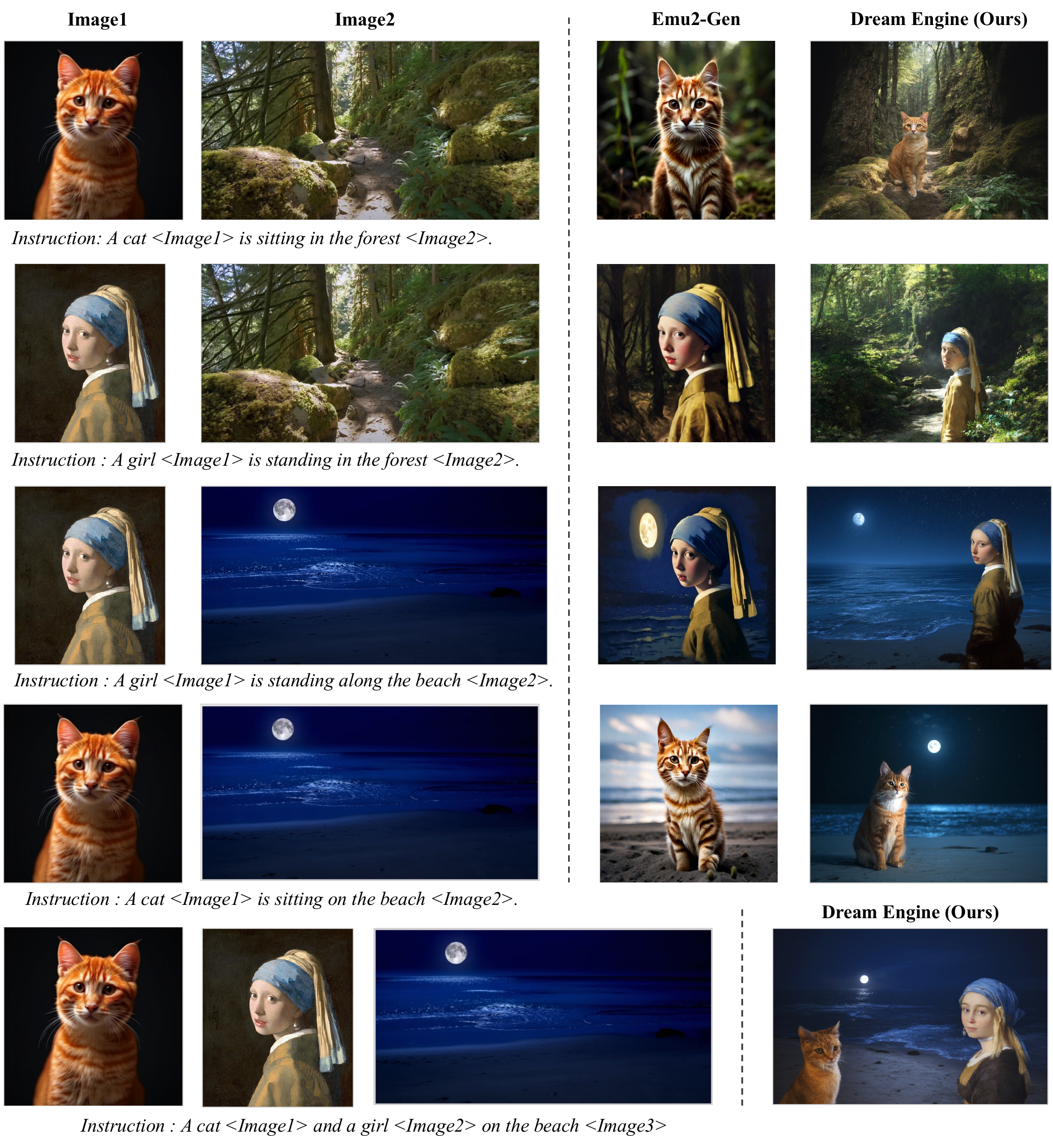}

\caption{Performance demonstration on Natural Object Background Merging where \model~ can understand complex text-image input. It can even set more than one object in the background (last line). }
\label{fig:example_front_background}
\end{figure*}

\paragraph{Multimodal Diffusion Transformer} MM-DiT structure~\citep{2024SD3}, is the basic structure for state-of-the-art text-to-image diffusion models such as SD3.5~\citep{2024SD3} and FLUX~\citep{flux}. It is built upon the Latent Diffusion Model (LDM)~\citep{ldm} and Diffusion Transformer (DiT)~\citep{DIT}. It concatenates textual conditioning information $c$ with noised latent embeddings $x$ into a unified sequence. Within the DiT module, MM-DiT employs distinct LayerNorm and MLP layers for each modality while merging their sequences during the attention mechanism. This design allows each representation to evolve within its own specialized space while still considering the influence of the other modality. We employ the embeddings of the timestep $t$ and pooled representation of text condition in the modulation mechanism of DiT. We use rectified flow matching~\citep{liu2022flowstraightfastlearning} as the training objectives to conduct text-to-image generation in Section~\ref{subsec:structure_model}.
 
\subsection{Model Design}
\label{subsec:structure_model}

We adopt the MM-DiT module from Stable Diffusion 3~\citep{2024SD3} model as the DiT model and Qwen2VL~\citep{Qwen2vl} as the LMM to compose \model.

\paragraph{Align LMM and MM-DiT}
As shown in Figure~\ref{fig:dream_engine}, we completely replace the text encoders including CLIP~\citep{radford2021clip} and T5~\citep{xue2021mt5} from the text-to-image diffusion models with the LMM to get a unified representation of text and image $c$. To align the representation space of pretrained LMM with that of previous encoders and enable the MM-DiT module to take image input, we add a straight-forward adapter layer consisting of a two-layer MLP. The adapter maps the output hidden states of LMM to the conditioning feature space of MM-DiT. We add the average pooling representations of the LMM condition and timestep embedding as the modulation embedding $y$ in the MM-DiT model. We remove the token length limit of original text encoders so that~\model~can take any sequence length of text-image input.
\paragraph{Blending Visual Feature for Better Consistency}


To control the visual consistency in image editing and objects-driven generation tasks, we add a skip connection for visual features in the LMM to avoid visual information loss in the LMM backbone model. As shown in Figure~\ref{fig:dream_engine}, the final hidden states of image patches $h_I$ are a weighted sum of the LLM output hidden states $h_{\mathrm{LLM}}$ and ViT image features $h_{\mathrm{ViT}}$:

\begin{equation}
    h_I = (1-r)\cdot h_{\mathrm{LLM}} + r\cdot h_{\mathrm{ViT}}
\end{equation}

The adjustable blending ratio \( r \) allows for the control of image feature consistency between the input and output images, tailored to specific applications. During the training phase, \( r\in[0,1] \) adheres to a uniform distribution, which enables the flexibility to assign various values to \( r \) during the inference process as shown in Figure~\ref{fig:vision_blending}.

\paragraph{Training Objectives} We adopt rectified flows~\citep{liu2022flowstraightfastlearning} to learn the transition between the target data distribution $\mathbf{x}_0$ and a standard normal distribution $\boldsymbol{\epsilon}$, i.e.

\begin{equation}
    \mathbf{z}_t = (1 - t)\cdot \mathbf{x}_0 + t\cdot \boldsymbol{\epsilon},
\end{equation}

where $t \in [0,1]$ represents the timestep, and $\mathbf{z}_t$ denotes the corresponding distribution at the $t$-th step. At each step, based on the current distribution $\mathbf{z}_t$, a condition $\mathbf{c}$, and the timestep $t$, the model directly parameterizes the velocity $v_\theta (\mathbf{z}_t, \mathbf{c}, t)$. This velocity is expected to approximate $\mathbf{x}_0 - \boldsymbol{\epsilon}$ during the flow matching process. It is important to note that the condition $\mathbf{c}$ can include interleaved text-image control, as opposed to solely text-based information as seen in the original text-to-image diffusion models. The training objective is to minimize the expected L2 loss by updating the model parameters $v_\theta$, with a weight $w_t$ assigned to each timestep, i.e.

\begin{equation}
    \min_{v_\theta} \int_{0}^{1} \mathbb{E} \left[ w_{t} \left\| (\mathbf{x}_0 - \boldsymbol{\epsilon}) - v_\theta (\mathbf{z}_t, \mathbf{c}, t) \right\|^2 \right] dt
\end{equation}

We use the Euler Discrete Scheduler~\citep{karras2022elucidatingdesignspacediffusionbased} following SD3~\citep{2024SD3} to set the timestep. The target data distribution $\mathbf{x}_0$ comes from the latent representation of VAE, which is the same as the VAE of SD3 following a standard Latent Diffusion Model~\citep{ldm} training process.

\subsection{Training Stages}
\label{subsec:training_stages}

\begin{figure*}[t]
\includegraphics[width=1\linewidth]{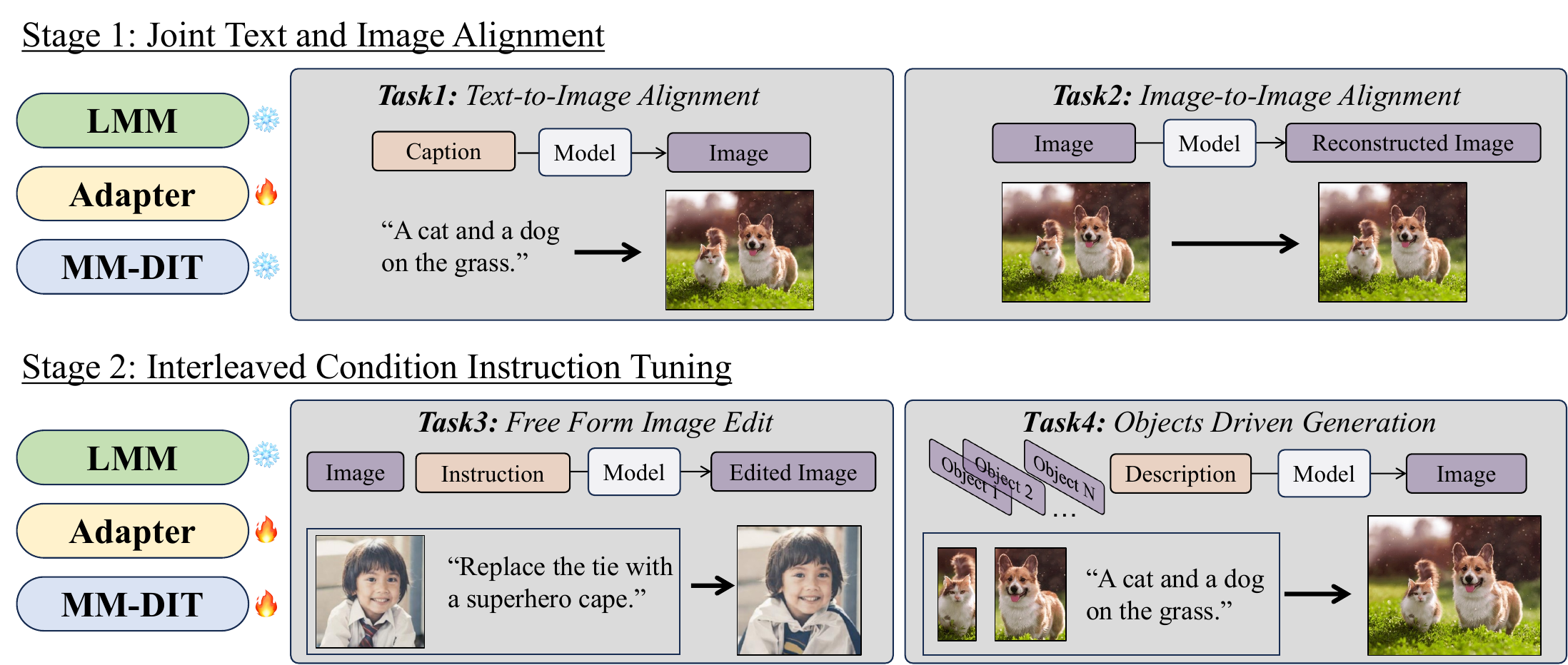}
\caption{Training stages and tasks of \model.}
\label{fig:stages}
\end{figure*}

Given that~\model~comprises two individually pretrained components—the LMM and the DiT—it is crucial to ensure their alignment. Adhering to the established practices outlined in the LMM literature~\citep{liu2023llava,QwenVL,internvl15,Cambrian-1}, we have structured the training process into distinct phases, each designed to unfreeze specific model components to promote stable and effective training. As shown in Figure~\ref{fig:stages}, our approach involves two primary training stages, where each stage has its own training tasks and trainable modules. In the S1 stage, we focus on training only the adapter layer, which facilitates the alignment of the representation spaces between the LMM and the DiT. During the S2 stage, we train both the adapter and the DiT, allowing for more sophisticated control over the generation process. We also show the training examples of each task in Figure~\ref{fig:stages}.


\paragraph{Stage 1: Joint Text and Image Alignment}

In the first stage, we focus on aligning the representation spaces of the LMM and DiT modules by training a dedicated adapter, while keeping the parameters of both the LMM and DiT frozen. This alignment process involves two complementary tasks. 

\begin{itemize}
    \item Task A: Text-to-Image Alignment. It leverages high-quality image-caption pairs to establish a foundational correspondence between textual descriptions and generated images, effectively replacing the original text encoders.
    \item  Task B: Image-to-Image Alignment. It is a self-supervised task that enables DiT to condition on image inputs. Specifically, DiT is trained to reconstruct input images based on the LMM’s image representations, thereby enhancing the consistency and fidelity of visual elements. 
\end{itemize}

Upon completing Stage 1 training, \model~acquires two core capabilities: \textbf{text-to-image generation} and \textbf{image variation}. Interestingly, we observe that the two tasks mutually reinforce each other. Even when trained solely on one task, the model demonstrates a certain degree of capability in the other task in a zero-shot manner. This finding suggests that the LMM inherently provides a unified representation space for text and images, which the DiT can effectively leverage during training. As shown in Table~\ref{tab:reconstruct}, our model trained without the Image-to-Image Alignment task can still achieve a relatively high (0.7+) CLIP score in the image reconstruction evaluation.

\paragraph{Stage 2: Interleaved Condition Instruction Tuning}

In the second stage, we unfreeze the DiT module and train it on two tasks that require interleaved image-text conditioning. 

\begin{itemize}
    \item Task C: Free-Form Image Editing. It takes an input image along with an editing instruction and outputs the edited image. We use the UltraEdit~\citep{ultraEdit} dataset as the dataset.
    \item Task D: Objects Driven Generation. It accepts multiple input images and a textual instruction, composing elements from the input images based on the given text to generate the output. For Task 4, we construct the training data using object detection datasets, such as COCO~\citep{lin2014mscoco}, pairing images with captions that describe the objects present.
\end{itemize}

After the second stage, the model gains the ability to handle interleaved image-text conditions during generation. Surprisingly, we observe emergent capabilities in \model. Notably, it can synthesize elements from different objects to generate cohesive images, as demonstrated in Figure~\ref{fig:example_object_mix}, despite such compositions not being explicitly present in the training data.

\begin{table*}[t]
    \centering
    \caption{\textbf{Performances on GenEval benchmark.} We split the methods to autoregressive and diffusion based models.
    }
    \resizebox{0.8\textwidth}{!}{
    \begin{tabular}{@{}cl
        >{\centering\arraybackslash}m{1.4cm}
        >{\centering\arraybackslash}m{1.4cm}
        >{\centering\arraybackslash}m{1.4cm}
        >{\centering\arraybackslash}m{1.4cm}
        >{\centering\arraybackslash}m{1.4cm}
        >{\centering\arraybackslash}m{1.4cm}
        >{\centering\arraybackslash}m{1.4cm}
        @{}}
        \toprule
         & \textbf{Method} & \textbf{Single Object} & \textbf{Two Object} & \textbf{Counting} & \textbf{Colors} & \textbf{Position} & \textbf{Attribute Binding} & \textbf{Overall} \\
        \midrule

        \multirow{6}{*}{\rotatebox{90}{\textit{Autoregressive}}}
        & Chameleon~\cite{2024Chameleon}  & - & - & - & - & - & - & $0.39$ \\
        & LWM~\cite{2024LWM}  & $0.93$ & $0.41$ & $0.46$ & $0.79$ & $0.09$ & $0.15$ & $0.47$ \\
        & LlamaGen~\cite{2024llamagen} &  $0.71$ & $0.34$ & $0.21$ & $0.58$ & $0.07$ & $0.04$ & $0.32$ \\
        & Show-o~\cite{2024Showo}  & $0.95$ & $0.52$ & $0.49$ & $0.82$ & $0.11$ & $0.28$ & $0.53$ \\
        & Emu$3$-Gen ~\cite{2024emu3} &  $0.98$ & $0.71$ & $0.34$ & $0.81$ & $0.17$ & $0.21$ & $0.54$ \\
        & Janus \cite{2024Janus}  & $0.97$ & $0.68$ & $0.30$ & \underline{0.84} & \textbf{0.46} & $0.42$ & $0.61$\\ 

        \midrule
        
        \multirow{12}{*}{\rotatebox{90}{\textit{Diffusion}}} 
        
        & LDM~\cite{2022LDM}  & $0.92$ & $0.29$ & $0.23$ & $0.70$ & $0.02$ & $0.05$ & $0.37$ \\
        & SDv$1.5$~\cite{2022LDM}  & $0.97$ & $0.38$ & $0.35$ & $0.76$ & $0.04$ & $0.06$ & $0.43$ \\
        & PixArt-$\alpha$~\cite{2023Pixelartalpha} & $0.98$ & $0.50$ & $0.44$ & $0.80$ & $0.08$ & $0.07$ & $0.48$ \\
        & SDv$2.1$~\cite{2022LDM} & $0.98$ & $0.51$ & $0.44$ & $0.85$ & $0.07$ & $0.17$ & $0.50$ \\
        & DALL-E $2$~\cite{2022DALLE2} &  $0.94$ & $0.66$ & $0.49$ & $0.77$ & $0.10$ & $0.19$ & $0.52$ \\

        & SDXL~\cite{2023SDXL} & $0.98$ & $0.74$ & $0.39$ & \textbf{0.85} & $0.15$ & $0.23$ & $0.55$ \\
        & IF-XL~\cite{2023IF}& $0.97$ & $0.74$ & $0.66$ & $0.81$ & $0.13$ & $0.35$ & $0.61$ \\
        & DALL-E $3$~\cite{2023dalle3}& $0.96$ & $0.87$ & $0.47$ & $0.83$ & \underline{0.43} & $0.45$ & $0.67$ \\
        &SDv3 Medium~\cite{2024SD3} & 0.98 & 0.74 & 0.63 & 0.67 & 0.34 &0.36 & 0.62
        \\
        &Flux.1 Dev~\citep{flux}& 0.98&	0.81	&
        \textbf{0.74}&	0.79	&0.22&	0.45 & 0.66 \\
        &SDv3.5 Large~\cite{2024SD3} & \underline{0.98} & \underline{0.89} & \underline{0.73} & 0.83 & 0.34 & \underline{0.47} &  \textbf{0.71}
        \\
        &\textbf{\model} & \textbf{1.00} & \textbf{0.94} & 0.64 & 0.81 & 0.27 & \textbf{0.49} & \underline{0.69}
        \\

        \bottomrule
    \end{tabular}}
    \label{tab:exp-geneval}
\end{table*}

\section{Experiments}
\label{sec:experiment}

\subsection{Dataset}

Table~\ref{tab:dataset} provides an overview of the datasets used for training~\model, along with the number of examples drawn from each source. In Stage 1, for the Text-to-Image Alignment task, we compile public image-caption datasets, including real-world images from CC12M~\citep{changpinyo2021cc12m} and model-generated images from JourneyDB~\citep{2024JDB}. Additionally, we synthesize a subset of high-quality images using diverse prompts with open text-to-image models, such as Flux.1 dev~\citep{flux} and Stable-Diffusion v3.5 Large~\citep{2024SD3}. For the Image-to-Image Alignment task, we rely solely on images from JourneyDB, as lower-aesthetic-quality images, such as images from CC12M, tend to degrade overall image reconstruction and text-to-image performance. In Stage 2, we utilize the UltraEdit~\citep{ultraEdit} dataset for the Free-Form Image Editing task and an internal object detection dataset for Object-Driven Generation. For the latter, we randomly select three objects from each image. Additionally, the name of each selected object must appear in the text caption to compose the conditioning input.

\begin{table}[t!]
\centering
\caption{Details on datasets used in training \model~ within the two training stages. }
\resizebox{0.47\textwidth}{!}{
\begin{tabular}{@{}cllc@{}}
\toprule
\textbf{Stage} & \textbf{Dataset} & \textbf{Task} & \textbf{Number} \\
\midrule
\multirow{4}{*}{1} & JourneyDB~\citep{2024JDB} & Text-to-Image Alignment & 4M \\
 & CC12M~\citep{changpinyo2021cc12m} & Text-to-Image Alignment & 4M \\
 & Synthetic Data & Text-to-Image Alignment & 4M \\
 & JourneyDB~\citep{2024JDB} & Image-to-Image Alignment & 4M \\
\midrule
\multirow{2}{*}{2}  & UltraEdit~\citep{ultraEdit} & Free Form Image Edit & 1M \\
 & Internal Data & Object-Driven Generation & 4M \\
\bottomrule
\end{tabular}}
\label{tab:dataset}
\end{table}

\subsection{Model and Training Details}

We initialize the LMM and DiT module of \model~from Qwen2VL-2B-Instruct~\citep{Qwen2vl} and Stable-Diffusion-3.5-Large~\citep{2024SD3}. The Adapter consists of a two-layer MLP with a middle projection dimension of 4,096 and uses SiLU as the activation function following the DiT module. In Stage 1, we freeze the parameters of the LMM and DiT modules and train the Adapter on the composed dataset for one epoch with a global batch size of 128. The learning rate is set to 1e-4, with 5\% warmup steps and a cosine learning rate scheduler. In Stage 2, we also fine-tune the DiT module using LoRA~\citep{hu2021lora} with a rank of 32 on all attention layers. The learning rate is set to 5e-5, while all other settings remain the same as in Stage 1. We do not fine-tune the LMM component of the model, thus preserving its original multimodal understanding capabilities. This design choice allows the model to be easily adapted into an omni-model, capable of performing both multimodal understanding and generation simultaneously. On the other hand, unfreezing the LMM during training has large potential in further improving the generation performance.

\subsection{Results and Comparisons}

\begin{table}[t]
\centering
\caption{Image reconstruction performance comparison on COCO and JourneyDB datasets.}
\resizebox{0.47\textwidth}{!}{
\begin{tabular}{@{}lcccc@{}}
\toprule
\multirow{2}{*}{\textbf{Method}} & \multicolumn{2}{c}{\textbf{COCO}} & \multicolumn{2}{c}{\textbf{JourneyDB}} \\
\cmidrule(lr){2-3} \cmidrule(lr){4-5}
 & CLIP~($\uparrow$) & L2~($\downarrow$) & CLIP~($\uparrow$) & L2~($\downarrow$) \\
\midrule
SeedTokenizer & 0.7760 & 0.5102 & 0.7921 & 0.5291 \\
EMU2-Gen & 0.8537 & \underline{0.3828} & \textbf{0.9299} & \underline{0.2869} \\
SEED-X & \underline{0.8595} & 0.4317 & 0.9017 & 0.4352 \\
\midrule
\model & \textbf{0.8714} & \textbf{0.2065} & \underline{0.9221} & \textbf{0.2052} \\
\quad - w/o I-to-I Alignment & 0.7184 & 0.6541 & 0.7536 & 0.6543 \\
\bottomrule
\end{tabular}}
\label{tab:reconstruct}
\end{table}

\begin{figure*}[t]
\vspace{5em}
\includegraphics[width=1\linewidth]{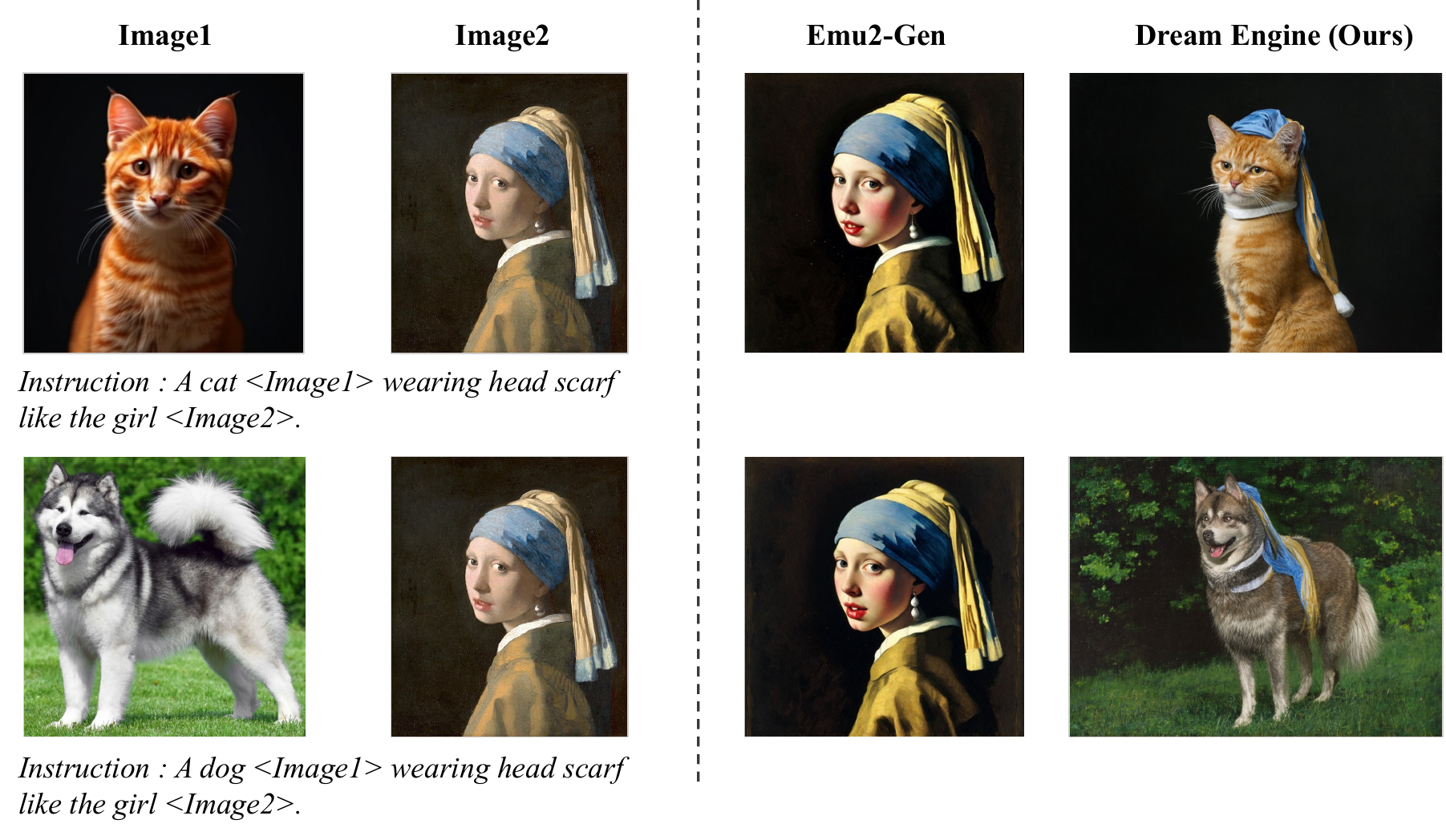}

\caption{ Performance demonstration on Object Driven Feature Mixing task. \model~ can understand the complex instruction while Emu2-Gen fails on the task.}
\label{fig:example_object_mix}
\end{figure*}

\begin{figure*}[t]
\centering
\includegraphics[width=1\linewidth]{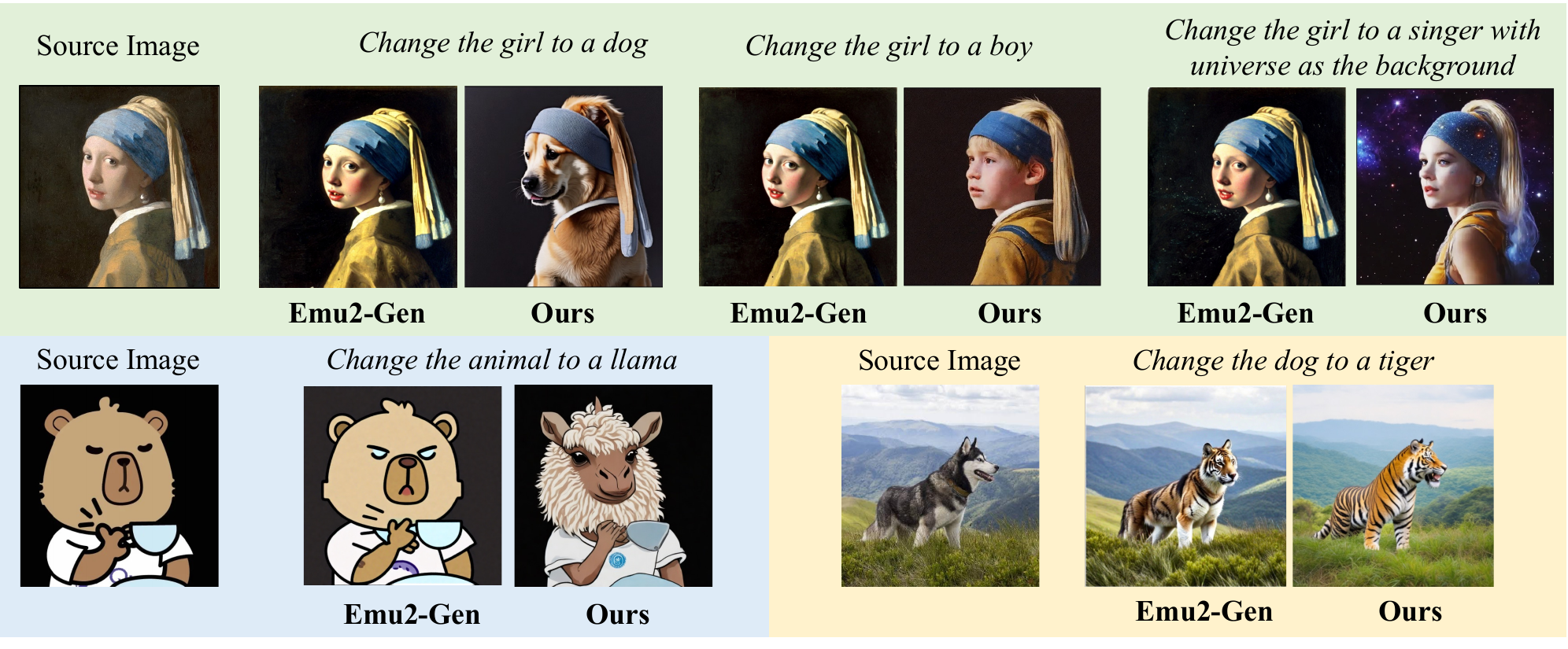}
\caption{Performance demonstration on Free Form Image Editing task. \model~ outperforms the counterpart Emu2-Gen model in both instruction following and output image quality. }
\label{fig:example_freeform_edit}
\end{figure*}

\paragraph{\mbox{Text-to-Image Generation}}

We evaluate the text-to-image generation capability of \model on the GenEval~\citep{ghosh2023genevalobjectfocusedframeworkevaluating} benchmark following Stage 1 training. The results, including fine-grained scores, are presented in Table~\ref{tab:exp-geneval}. Built upon the SDv3.5~\citep{2024SD3} model, \model achieves a competitive overall score of 0.69, closely matching the original model’s 0.71 despite excluding its native text encoders. Moreover, \model outperforms all other counterparts, demonstrating the effectiveness and efficiency of our text-to-image alignment training in preserving instruction-following capabilities while replacing the original text encoders of diffusion models to enable more complex interleaved conditions.

\begin{figure*}[t]
\centering
\includegraphics[width=1\linewidth]{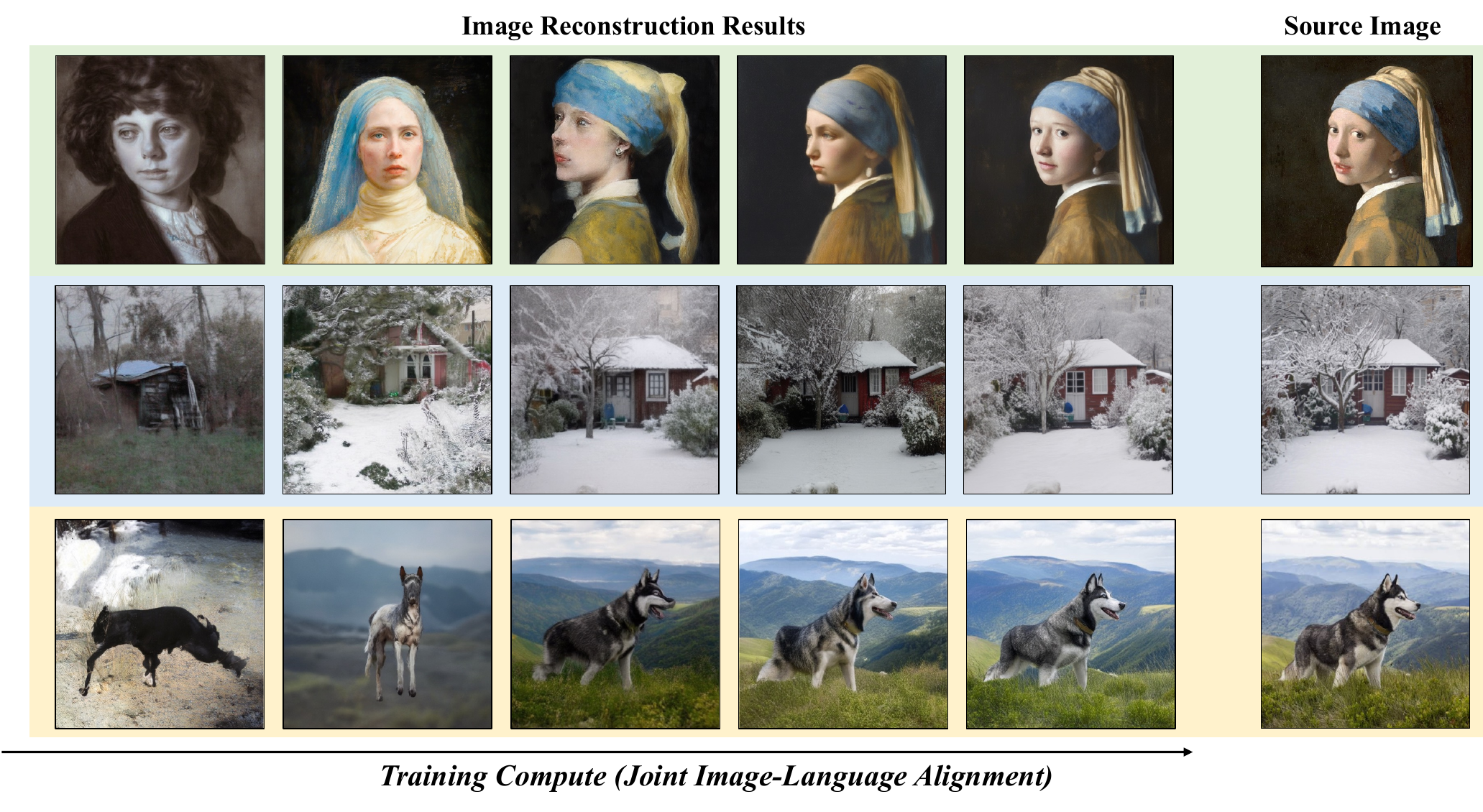}
\caption{Image reconstruction performance dynamics during training. We can see that there is a concept-to-detail transition during the training period.}
\label{fig:recon}
\end{figure*}

\paragraph{Image Reconstruction}
We introduce an Image Reconstruction Benchmark to evaluate the preservation of visual features in our Image-to-Image alignment task during Stage 1. This capability is essential for generating images conditioned on input images. To construct the benchmark, we randomly sample 100 images from the JourneyDB development set and 100 images from the COCO development set. We assess the similarity between the original and reconstructed images using the CLIP~\citep{radford2021clip} score and L2-Distance. As shown in Table~\ref{tab:reconstruct}, we compare the performance of \model~ against several baselines with similar architectures, including SeedTokenizer~\citep{seed-tokenizer}, EMU-2~\citep{emu2}, and SeedX~\citep{2024SeedX}, which also integrate LMMs and diffusion models for generation. The results demonstrate that our model achieves the best average image reconstruction performance across both subsets of the benchmark. Notably, it achieves outstanding performance on the L2 distance metric, which emphasizes pixel-level consistency, surpassing the second-best model by 46\% on the COCO subset and 28\% on the JourneyDB subset.

\paragraph{Generation with Text-Image Interleaved Control}

After Stage 2 training, \model~acquires the capability to incorporate text-image interleaved control during the image generation process. We showcase several applications of the model in this paper and compare its performance with Emu-2~\citep{emu2}, the most relevant baseline that also supports text-image interleaved control.

\begin{figure*}[t]
\centering
\includegraphics[width=1\linewidth]{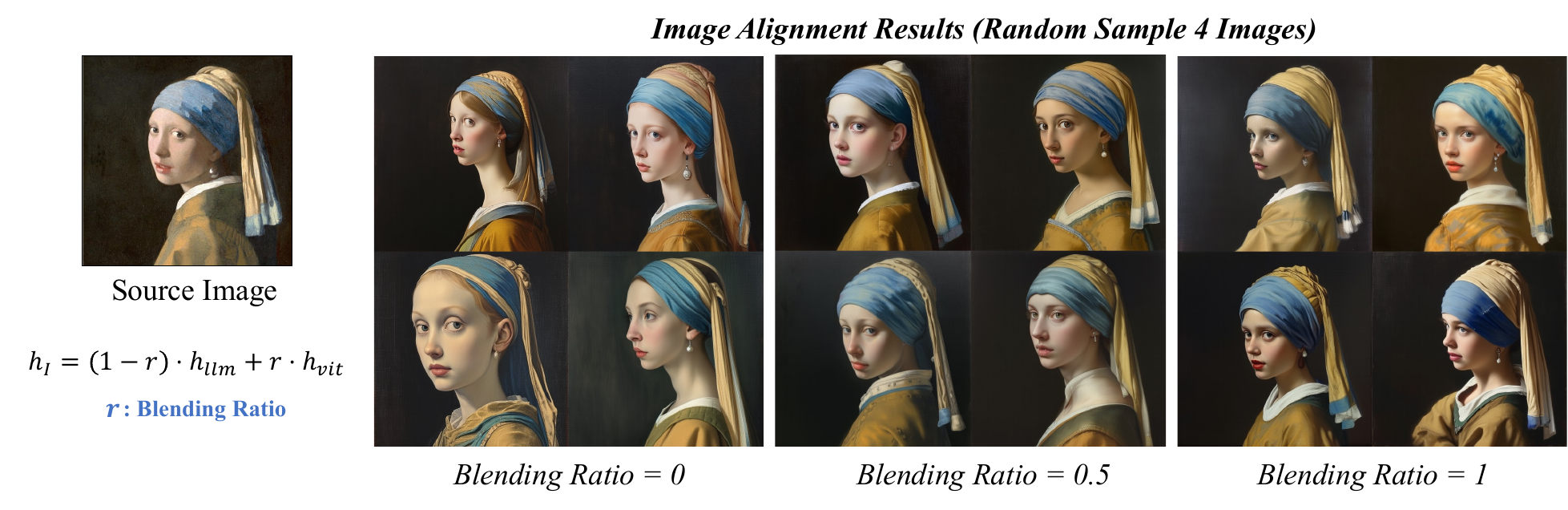}
\caption{The ablation on visual blending ratio in the ViT module. It reveals that a higher blending ratio results in greater consistency during image reconstruction tasks.}
\label{fig:vision_blending}
\end{figure*}

\begin{itemize}
    \item[1.] \textbf{Natural Object Background Merging:} Figure~\ref{fig:example_front_background} illustrates an example application where objects are merged into different backgrounds based on a provided hint image. The results demonstrate that \model can seamlessly place the main object into various backgrounds in a more natural manner even when there are multiple objects, rather than merely copying and pasting the objects.
    \item[2.] \textbf{Object Driven Feature Mixing:} Figure~\ref{fig:example_object_mix} demonstrates an emergent ability of \model~to generate images by combining visual features from given images based on text instructions—an area where the EMU-2 model fails to follow instructions accurately. Notably, these examples are not present in the training dataset, which was constructed directly from an object detection dataset. This highlights the significant potential of LMMs as unified multimodal instruction encoders for image generation. \model~ effectively decouples complex elements from different images using simple text prompts and produces a unified representation, showcasing its versatility and robustness.
    \item[3.] \textbf{Free Form Image Editing:} Figure~\ref{fig:example_freeform_edit} presents the results of our free-form image editing task. \model~consistently demonstrates superior ability to follow edit instructions compared to EMU-2. Notably, \model~can handle complex editing instructions, such as simultaneously modifying both the object and background. This further highlights the effectiveness of LMMs in providing a unified representation space that seamlessly integrates image and text conditions.
\end{itemize}

\subsection{Discussions}

\paragraph{Understand the training dynamics of \model}

To better understand how \model~leverages an LMM and a text-to-image model to achieve complex text-image instruction following ability during the training process, we examine the image reconstruction results at different training stages in Stage-1. The results are shown in Figure~\ref{fig:recon}. We observe a clear concept-to-detail progression during the training of \model. For example, as illustrated in the first two columns of Figure~\ref{fig:recon}, the model initially reconstructs the primary concepts in the images, such as ``girl,'' ``house with snow,'' and ``dog'' in the given examples. 

In the later stages, the model begins to learn to reconstruct more fine-grained details, such as colors, shapes, and poses. We believe this unique training dynamic stems from the nature of the LMM, which provides a unified representation space where images and text are well aligned. Thus, at the beginning of training, even if the text-to-image diffusion model has not yet seen image conditions, the image representations provided by the LMM are aligned with text features. This alignment enables the model to generate conceptually aligned images through the bridge of text.

\paragraph{Ablation on Balancing Visual Consistency}

We conduct an ablation study on the Blending Visual Feature mechanism within the Vision Transformer (ViT) module. As shown in Figure~\ref{fig:vision_blending}, varying the blending ratio in the image-image alignment task produces notably different results. Higher blending ratios lead to greater consistency in image reconstruction, while lower ratios introduce more variation in the output. This mechanism offers flexible control over object consistency, benefiting various downstream tasks such as image editing and object-driven feature mixing.

\section{Related Work}
\label{sec:related}
\subsection{Image Generation with Complex Control}
Recent progress in controlled image generation using diffusion models has been significant. Researchers have explored various conditioning strategies—ranging from low-level cues like canny edges and depth maps~\citep{ye2023ip-adapter, controlnet} to higher-level guidance provided by reference images~\citep{SDEdit}—to steer the generative process. For instance, methods such as IP-Adapter~\citep{ye2023ip-adapter} and ControlNet~\citep{controlnet} incorporate additional control signals into standard text-to-image frameworks, thereby allowing more precise manipulation of generated content.
In parallel, several works have leveraged visual elements from input images to further guide the generation process. DreamBooth~\citep{ruiz2023dreamboothfinetuningtexttoimage} and Textual Inversion~\citep{gal2022imageworthwordpersonalizing}, for example, adopt optimization-based approaches to adapt models to specific reference images. Although effective, these methods typically require extensive fine-tuning for each new input, limiting their practicality. To address these limitations, approaches like SuTI~\citep{suti} and Subject-diffusion~\citep{ma2024subjectdiffusionopendomainpersonalizedtexttoimage} have aimed to scale the fine-tuning process so that models can generalize across diverse reference images. However, these strategies still tend to be both time- and resource-intensive, highlighting the ongoing need for more efficient mechanisms for image generation with complex controls.

\subsection{Connecting LMMs with Diffusion Models}

Recent studies integrate LMMs with diffusion generators, leveraging the strengths of both paradigms. 
One straightforward approach employs LMMs to interpret complex text-image conditions and generate pure textual representations, which then guide image generation models~\citep{Mini-Gemini}.
Moreover, Seed-Tokenizer~\citep{seed-tokenizer} expands the LMM vocabulary by introducing discrete vision tokens that serve as robust conditioning signals for diffusion models, while Seed-LLama~\citep{seedllama} pre-trains a discrete image tokenizer that decodes visual codes into realistic images using pretrained diffusion models. Similarly, M-VADER~\citep{MVADER} aligns semantic consistency between language models and diffusion decoders through training on extensive image-text pair datasets. Methods such as GILL~\citep{koh2023GILL}, MiniGPT5~\citep{zheng2023minigpt5} Emu~\citep{sun2023emu1} further advance this integration by mapping the embedding spaces of language models to diffusion models, and NExT-GPT~\citep{nextgpt} and Any-GPT~\citep{Zhan2024AnyGPT} even broadens the scope to include modalities like audio and video. Additionally, DreamLLM~\citep{dreamllm}  employs a novel strategy by transferring differential gradients from image diffusion models to language models, thereby enabling free-form, interleaved content generation.
To enhance flexible control in image generation, BLIP-Diffusion~\citep{li2023blipdiffusionpretrainedsubjectrepresentation} leverages LMMs to jointly encode image and text inputs, projecting them into the text conditioning space of diffusion models to better handle complex instructions. 
Moreover, Kosmos-g~\citep{Kosmos-G} and Emu-2~\citep{emu2} explore the multimodal in-context control for image generation. 
Seed-X~\citep{2024SeedX} using a similar architecture to model multi-granularity visual semantics for better generation in the real world applications.
While promising, these approaches mainly extend text-to-image generation and often fall short in improving image quality or managing compositional tasks with arbitrary text-image interleaved control, limiting their real-world applicability.

\section{Conclusion}

In this work, we introduced \model, a novel framework that enables sophisticated text-image interleaved control without complex architectural modifications. Our method bridges Large Multimodal Models and text-to-image diffusion models through a lightweight projector layer and efficient two-stage training paradigm. It demonstrates superior capabilities in handling multiple image inputs and compositional instructions while also achieving competitive performance on the GenEval benchmark (0.69). The success of \model~demonstrates that LMMs can effectively replace traditional text encoders in text-to-image diffusion models while expanding their capabilities to include sophisticated multimodal control. 

Looking ahead, our work opens up new possibilities for creative image generation applications where users can precisely orchestrate visual compositions through natural language instructions and multiple reference images. Future research directions could explore extending this framework to other modalities, such as video or 3D content, and investigating ways to further enhance the semantic understanding of complex multimodal instructions.

{
    \small
    \bibliographystyle{ieeenat_fullname}
    \bibliography{main}
}


\end{document}